\pdfoutput=1

\documentclass[11pt]{article}

\usepackage{authblk}

\usepackage{acl}

\usepackage{times}
\usepackage{latexsym}

\usepackage[T1]{fontenc}

\usepackage[utf8]{inputenc}

\usepackage{enumitem}
\usepackage{microtype}
\usepackage{amsmath}
\usepackage{amssymb, bm}
\usepackage{multirow}
\usepackage{graphicx}
\graphicspath{ {./images/} }
\usepackage{subfig}
\usepackage[vlined, linesnumbered]{algorithm2e}
\RestyleAlgo{ruled}

\usepackage{xcolor}

\usepackage{dblfloatfix}

\title{Seeded Hierarchical Clustering for Expert-Crafted Taxonomies}

\author[1]{Anish Saha}
\author[1]{Amith Ananthram}
\author[1]{Emily Allaway}
\author[2]{Heng Ji}
\author[1]{Kathleen McKeown}
\affil[1]{Columbia University}
\affil[2]{University of Illinois Urbana-Champaign}
\affil[ ]{\texttt{
\{anish.s,amith.ananthram\}@columbia.edu
}}
\affil[ ]{\texttt{
\{eallaway,kathy\}@cs.columbia.edu,
hengji@illinois.edu
}}

\begin{document}

\maketitle
\begin{abstract}
Practitioners from many 
disciplines (e.g., political science) use expert-crafted taxonomies to make sense of large, unlabeled corpora. 
In this work, we study Seeded Hierarchical Clustering (SHC): the task of automatically fitting unlabeled data to such taxonomies using a small set of labeled examples. 
We propose \textsc{HierSeed}, a novel weakly supervised 
algorithm for this task that uses only a small set of labeled seed examples in a computation and data efficient manner. 
\textsc{HierSeed} assigns documents to topics by weighing document density against topic hierarchical structure. 
It outperforms unsupervised and supervised baselines for the SHC task on three real-world datasets.
\end{abstract}
\vspace{-10pt}
\begin{figure*}[tp]
\centering
\hfill
\includegraphics[width=\textwidth]{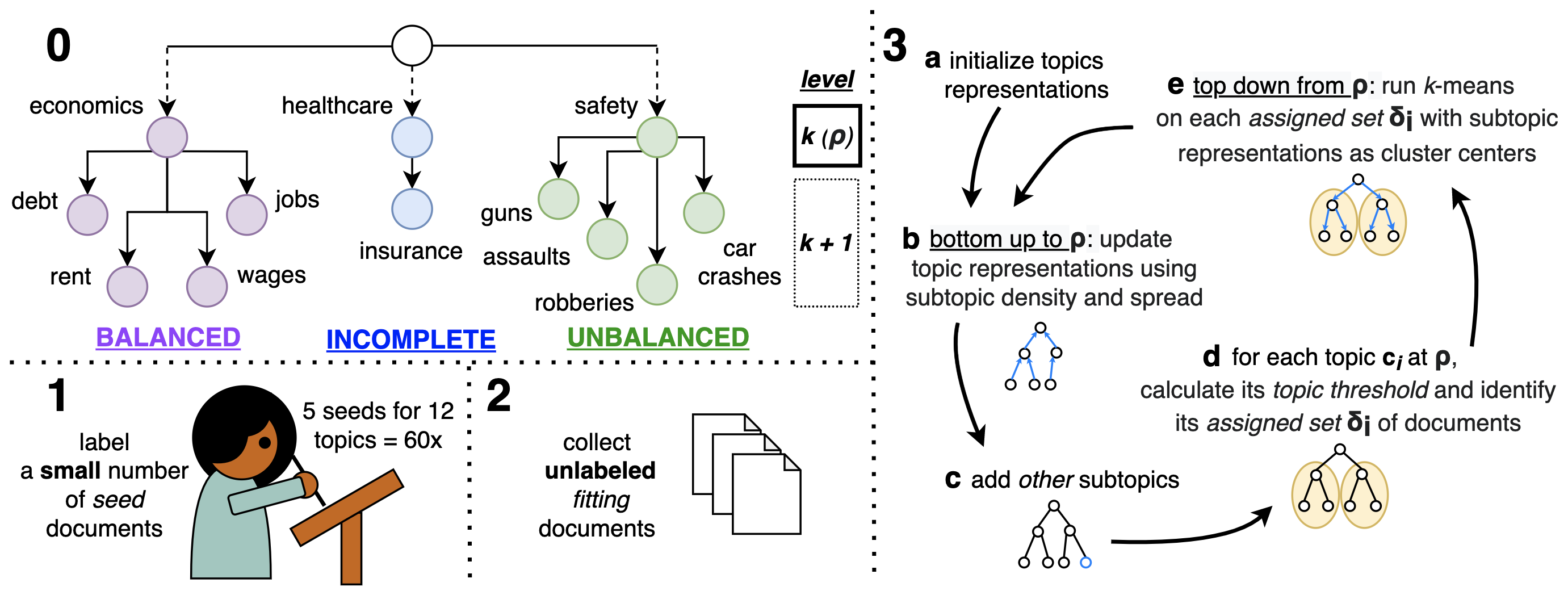}
\caption{A researcher wants to track public well-being using a large, unlabeled social media corpus.  She creates a taxonomy of relevant topics (\textbf{0}) -- it does not cover every document in her dataset.  Moreover, as it is hand-crafted, it is \textit{unbalanced} and \textit{incomplete}.  She can't annotate a large number of examples.  With only a few \textit{labeled seed} examples for each topic (\textbf{1}) and her large \textit{unlabeled fitting} set (\textbf{2}), \textsc{HierSeed} efficiently identifies the documents related to every topic via an iterative discriminative algorithm that balances their density against their spread (\textbf{3}).}
\label{tab:example}
\end{figure*}

\section{Introduction}
Practitioners across a diverse set of domains that include web mining, political science and social network analysis rely on machine learning techniques to understand large, unlabeled corpora \citep{alfonseca2012pattern,grimmer2010bayesian,yin2014dirichlet}. 
In particular, they often need to fit this data to taxonomies (i.e., hierarchies) constructed by non-technical domain experts using only a few labeled examples. In this work, we formalize this task, \textbf{Seeded Hierarchical Clustering (SHC)},
and propose a novel algorithm, \textbf{\textsc{HierSeed}}, 
for it.

Consider a researcher analyzing social media to track public feeling around a hierarchy of well-being indicators (see Figure \ref{tab:example}). 
Working with such taxonomies can be challenging.  Since they are hand-crafted by domain experts to explore a particular area of focus, they may be \textbf{unbalanced} (with subtopics that over-represent one aspect of their parent topic) or \textbf{incomplete} (with subtopics that are only partially enumerated). Moreover, these hierarchies may not fully explain every document in large, diverse corpora. Finally, given their domain specificity, producing 
many labeled examples for each topic in such taxonomies can be expensive.  
\textbf{SHC} incorporates these challenges as constraints:
given only a user-defined topic hierarchy and a few labeled examples, the task is to
assign documents from a much larger corpus to the individual topics.

While many unsupervised techniques and their hierarchical extensions
automatically discover latent structure within text corpora, they are difficult to 
integrate with user-defined taxonomies (e.g. \citet{blei2003latent,lloyd1982least,campello2013density}).  
Moreover, as these methods often rely on centroids, density metrics and maximum likelihood objectives to discover dataset partitions, 
they may produce clusters that
favor the \textit{denser, semantically more coherent} regions of an unbalanced taxonomy at the expense of the \textit{sparser but more diverse} regions. 
Although supervised hierarchical methods avoid these issues,
they are usually very data intensive.

To address these challenges, we propose \textsc{HierSeed}, a weakly supervised hierarchical method for fitting large unlabeled corpus to a user-defined taxonomy.
It assigns documents to topics by weighing document density against a topic's local hierarchical structure. 
To accommodate imbalance or incompleteness, \textsc{HierSeed} constructs and uses topic representations that account for
subtopic \textit{density} (degree of semantic coherence) and \textit{spread} (degree of semantic divergence) around each topic.  
As it uses only a few labeled \textit{seed} examples to optimize  
its objective in a non-parametric fashion, it is both data and computationally efficient.
We evaluate \textsc{HierSeed} on three real-world newswire and scientific datasets
and show that it outperforms state-of-the-art unsupervised and supervised baselines on this new, difficult task.

Our contributions are:
(1) we \textbf{formalize the task} of \textbf{Seeded Hierarchical Clustering}, 
(2) we present \textsc{HierSeed}, a 
novel algorithm that uses only a few labeled examples to efficiently fit a large corpus to a user-defined (\textbf{unbalanced} or \textbf{incomplete}) hierarchy
and (3) we show it \textbf{outperforms} existing 
methods on three real-world datasets from different domains. An implementation of our algorithm and its evaluation will be made available.

\section{Related Work}
Unsupervised methods like LDA and K-Means \citep{blei2003latent,lloyd1982least,macqueen1967some} are flat clustering techniques that 
have been successfully extended to hierarchies \citep{griffiths2003hierarchical,isonuma2020tree}. 
While both we and \citet{chen2005novel} apply K-Means iteratively, they rely on hierarchical clustering to discover the number of topics at each level.
None of these methods 
can detect a user-defined hierarchy. There is work on  taxonomy construction and expansion \citep{hearst1992automatic,wang2007language,shen2018hiexpan} though it cannot be used for document assignment.

Supervised hierarchical classification techniques can be categorized into flat, local and global approaches \citep{silla2011survey}. 
Flat approaches \citep{hayete2005gotrees, barbedo2006automatic}
ignore the hierarchy, whereas local approaches \citep{koller1997hierarchically,shimura2018hft} rely on multiple local classifiers, propagating errors down the hierarchy. 
Global approaches \citep{zhou2020hierarchy,huang2021hierarchy} encode the entire hierarchy and predict all labels at once. 
This leads to better performance and has
become common for this task.
Nevertheless, these approaches require a large number of labeled training examples for good performance, making them difficult to use in data-scarce scenarios.

Semi-supervised hierarchical
methods require much less labeled data as they make use of topic labels \citep{mao2012sshlda,gallagher2017anchored}.
JoSH \citep{meng2020hierarchical}
uses a tree and text embedding model to 
jointly embed a
taxonomy and a corpus 
in a spherical space, using category names to mine 
words relevant to each topic. 
Weakly-supervised classifiers like HClass \citep{meng2019weakly} leverage provided keywords and documents for each topic to generate a set of pseudo documents for pretraining, then self-train on unlabeled data. Despite their strengths, these methods require more labeled data and more expensive finetuning than \textsc{HierSeed}.  Moreover, they work best on taxonomies with classes describable by simple labels.

In contrast to this prior work, \textsc{HierSeed} takes any taxonomy (unbalanced or incomplete) and a few labeled seeds and learns a discriminative hierarchical representation that allows assigning relevant documents to each of its component topics.

\section{Methodology}
We propose \textsc{HierSeed}, a weakly-supervised algorithm for \textbf{Seeded Hierarchical Clustering} that uses dense embeddings and 
the structure of the embedding space
to represent topics in the same embedding space (\S\ref{sec:defs}).
We initialize the representation for each topic using its seed documents (\S\ref{sec:taxoinit}) and then update the representation in a bottom-up manner by considering both a topic's children and the density of documents around it (\S\ref{sec:taxorep}). 
Finally,
we balance the hierarchy and assign documents to each topic in a top-down manner (while also updating its representation). 
We repeat this iteratively until convergence (see Algorithm \ref{alg:shc}, Figure \ref{tab:example} (3)).

\begin{algorithm}[t]
    \caption{\textsc{HierSeed}}\label{alg:shc}
    \KwIn{
    A corpus $\mathcal{D}$; 
    seeds $\mathcal{S}$ and taxonomy $\mathcal{T}$ of height $N$; 
    pivot level $\rho$.
    }
    \KwOut{
    Learned topic representations for all $c_i \in \mathcal{T}$; 
    set of relevant documents $\delta_i \subset \mathcal{D}$ for each $c_i$.
    }
    Initialize topics $c_i \in \mathcal{T}$ with mean of seed documents $\mathcal{S}_i$ embeddings \\
    \While{Equation \ref{eq:objectivefn} is minimized}{
     \textbf{M-bottom-up}, $l$ from $N-1$ to $\rho$: \\
        \quad update $c_i \in C^l$ with Eq.~\ref{eq:weightmeanupdaterep} \\
    \textbf{Extend taxonomy}, $\forall c_i$
     \textbf{Add ``other''} topic to 
     $\mathrm{ch}(c_i)$
     (Eq.~\ref{eq:othercatequation})\\
     \textbf{E-at-level-}$\rho$, for each $c_i \in C^{\rho}$:\\
     \quad calc. topic threshold $\tau(c_i)$
     (\S\ref{sec:topicthreshass})\\
     \quad identify documents $\delta_i$ (Eq. \ref{eq:coreassignment})\\
     \textbf{M-top-down}, $l$ from $\rho$ to $N$, $c_i \in C^l$: \\ 
     \quad run K-means on $\delta_i$, $k=|\mathrm{ch}(c_i)|$ \\ 
      \quad set $\mathrm{ch}(c_i)$ to K-means cluster centers\\
     \textbf{E-top-down}, $l$ from $\rho$ to $N$, $c_i \in C^l$: \\ 
     \quad set $\delta_i$ to K-means clusters\\
    }
\end{algorithm}

\subsection{Definitions}
\label{sec:defs}

\paragraph{Problem Formulation}
Given an unlabeled corpus $\mathcal{D}$ (the \textit{fitting set}), a hierarchy of topics $\mathcal{T}$\footnote{We assume the hierarchy is relevant to the corpus.} of height $N$ 
and a \textit{seed} documents set $\mathcal{S}$ for each topic, 
the aim of \textbf{Seeded Hierarchical Clustering} 
is to assign 
documents to their relevant topics in $\mathcal{T}$.

Let $d_{i} \in \mathcal{D}$ be unlabeled \textit{fitting} documents, $c_i \in \mathcal{T}$ topics, and
let $\mathcal{S}_i$ be a set of \textit{seed} documents for topic $c_i$. 
Note that $\mathcal{S}_i$ may or may not be a subset of the corpus $\mathcal{D}$.
The aim is to find the set of documents $\delta_i \subset \mathcal{D}$ most relevant to each $c_i$.
Here, a document may belong to multiple topics. 

Note that we use $C^l$ to denote all topics at level $l$.
We denote children of a topic $c_i$ as  $c_j^{(i)} \in \mathrm{ch}(c_i)$.

\begin{figure}[tp]
\centering
\includegraphics[width=\linewidth]{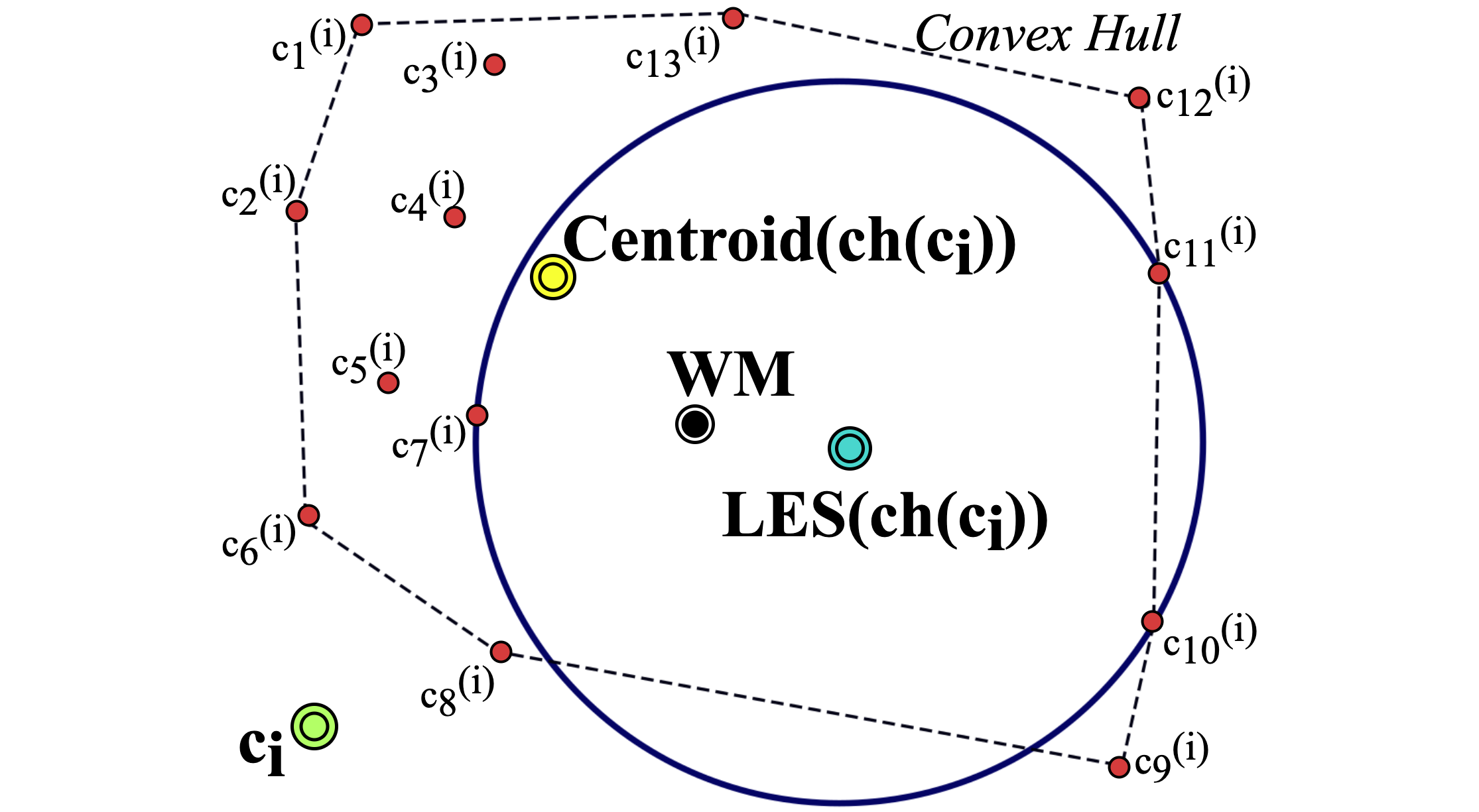}
\caption{The topic $c_i$, its children $c_j^{(i)} \in \mathrm{ch}(c_i)$, their centroid $\mathrm{centroid}(\mathrm{ch}(c_i))$, the center of $\mathrm{LES}\big(\mathrm{ch}(c_i)\big)$ and their weighted-mean $\mathrm{WM}$.}
\label{tab:LES}
\end{figure}

\paragraph{Background}
The \textbf{Largest Empty Sphere (LES)}~\citep{schuster2008largest}
on a set of points $\mathcal{P}$, is the largest $d$-dimensional hypersphere containing no points from $\mathcal{P}$ but centered within its convex hull. In \textsc{HierSeed}, for each topic $c_i$, we calculate $\mathrm{LES}(\mathrm{ch}(c_i))$, the center of the LES on $c_i$'s subtopics (i.e., $\mathrm{ch}(c_i)$) (see Figure~\ref{tab:LES}).

$\mathrm{LES}(\mathrm{ch}(c_i))$ has a particularly desirable property.  Since it is as far as possible from all of its subtopics, but not too far from any particular subtopic, while also lying inside the subtopic convex hull, it helps ensure a more evenly spread surrounding document density.  The main topic is adequately desensitized to any particular subtopic cluster.
Recall the well-being taxonomy from Figure \ref{tab:example}. The \textit{safety} subtree is unbalanced -- $3$ of its subtopics are semantically related (\textit{guns}, \textit{assaults}, \textit{robberies}).  
Using the centroid to represent \textit{safety} would therefore overly favor documents related to violence at the expense of documents related to \textit{car crashes} (an enumerated subtopic) or \textit{workplace accidents} (an unenumerated but relevant subtopic).  As  $\mathrm{LES}(\mathrm{ch}(\textit{safety}))$ is informed by its subtopic spread, it is less sensitive to this imbalance.

\subsection{Topic Initialization}
\label{sec:taxoinit}

We obtain document representations by passing each document through a word-embedding model.
The representation of each topic $c_i$ 
is initialized as the mean of the embeddings of all seed documents corresponding to that topic.

Let the level of a topic be $\lambda(c_i)$\footnote{As is standard, $\lambda$ increases as we move down the tree.}. 
We choose a \textbf{pivot level} $\rho$, in a hierarchy, such that our algorithm computes representations and performs clustering only for topics at the pivot level or below (i.e., $\lambda(c_i) \ge \rho$).
This hyperparameter may be set experimentally or through domain knowledge.

The pivot is useful because user generated hierarchies tend to become imbalanced or incomplete at lower levels. 
Therefore, $\rho$ lets us choose an intermediate level such that all topics with $\lambda(c_i) < \rho$ are considered to be ``complete" or fully represented by their children and can be derived without seeds.
For example, in the 
taxonomy in Figure \ref{tab:example}, the topics above level $k$ are well defined.
However, down the hierarchy, the topics start getting \textit{sparse} or \textit{unbalanced}.
Thus, level $k$ serves as a good $\rho$.

As a result, our proposed algorithm \textit{only} uses seed documents for topics with levels $\lambda(c_i) \ge \rho$, further reducing the labeled data required.

\subsection{Learning Topic Representations}
\label{sec:taxorep}
Generally, topics lower in the hierarchy are specific and fine-grained with cohesive seeds. 
In contrast, top level topics are coarser with seeds that are often scattered.
Thus, we need to obtain better representation at these top levels while ensuring representativeness of descendent topics.

To counter this imbalance, we update a non-leaf topic in a bottom-up fashion as a function of: itself, its children and their "spread". 
Let $C^l$ be all topics at level $l$.
Then, for each level $l$ from 
the penultimate level to $\rho$, for each $c_i \in C^l$, update:
\begin{equation}
\label{eq:weightmeanupdaterep}
\begin{aligned}
    c_i = \mathrm{WM}\Big(
        \{c_i\} \:\cup \mathrm{ch}(c_i) \:\cup \{\mathrm{LES}\big(\mathrm{ch}(c_i)\big)\} 
    \Big)
\end{aligned}
\end{equation}
Here, $\mathrm{WM}$ is the weighted-mean, $\mathrm{LES}\big(\mathrm{ch}(c_i)\big)$ is center of the \textit{Largest Empty Sphere} formed by the children (\S\ref{sec:defs}) and $\mathrm{ch}(c_i)$ is the set of children of $c_i$. Since the dimension of our embedding space $d$ is large compared to the number of child topics, computing the $\mathrm{LES}$ is intractable. Therefore, 
we propose an approximate method to estimate the center of the $\mathrm{LES}$ (see Appendix~\ref{sec:appendixLES}).

The weights in the $\mathrm{WM}$ are hyperparameters. When the weights of all terms in $\mathrm{ch}(c_i)$ are set to $\frac{1}{|\mathrm{ch}(c_i)|}$, we get their centroid (Figure \ref{tab:LES}). 
This is a good default setting as the updated representation of $c_i$ is closer to the centroid of its sub-topics.

Moving toward the centroid, however, may end up favoring topics that happen to be close to each other or denser. 
By taking a weighted-mean of $c_i$, the centroid of its subtopics and the $\mathrm{LES}$ of its subtopics, 
we balance finding a space that is relatively empty ($\mathrm{LES}$) against a space that is relatively dense (centroid) (see Figure~\ref{tab:LES}). This produces topic representations robust to hierarchy imbalances.

\begin{figure*}[tp]
\centering
\hfill
\subfloat[\centering Extending with "Other" Category.]{{\includegraphics[width=5.5cm]{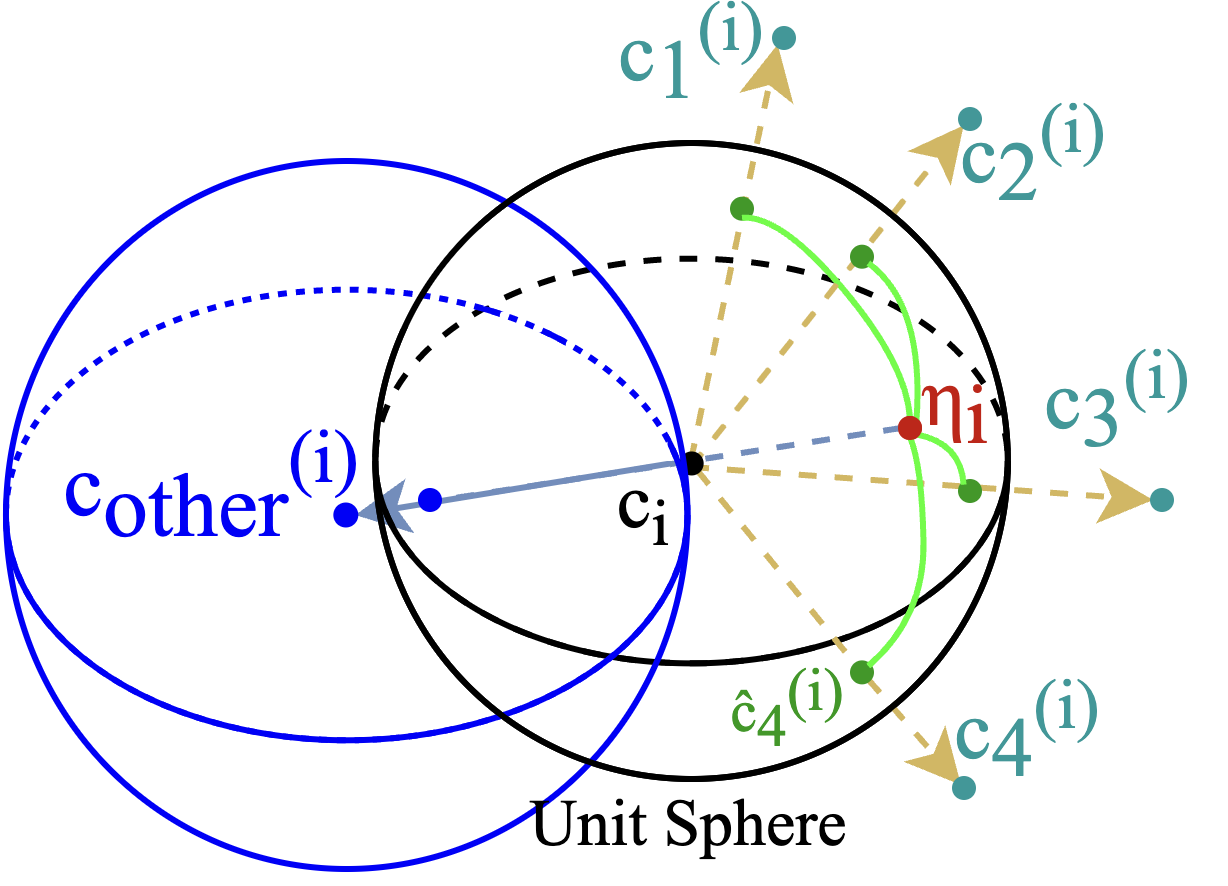} }}
\hfill
\subfloat[\centering Topic with children]{{\includegraphics[width=4.5cm]{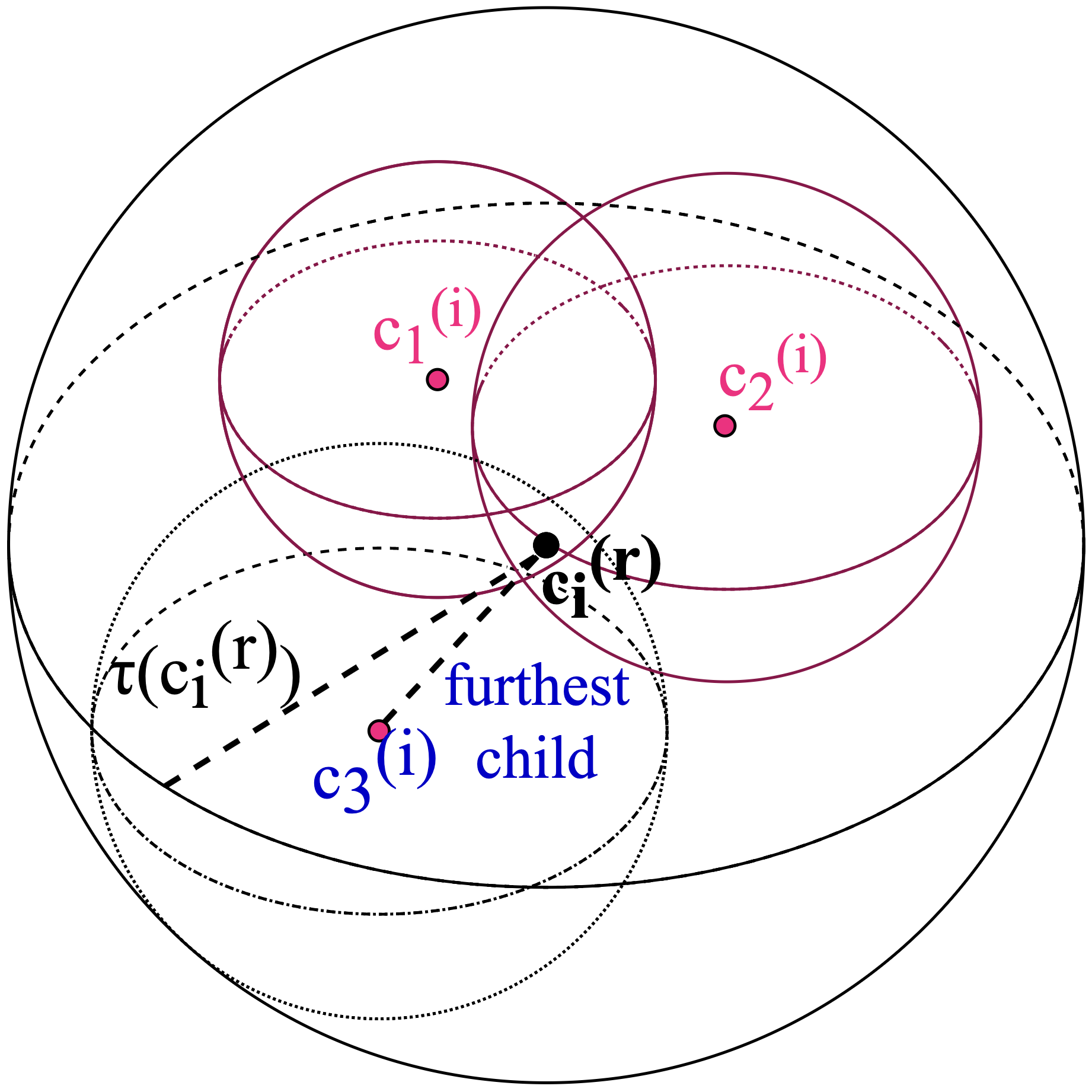} }}
\hfill
\subfloat[\centering Topic without children]{{\includegraphics[width=5.5cm]{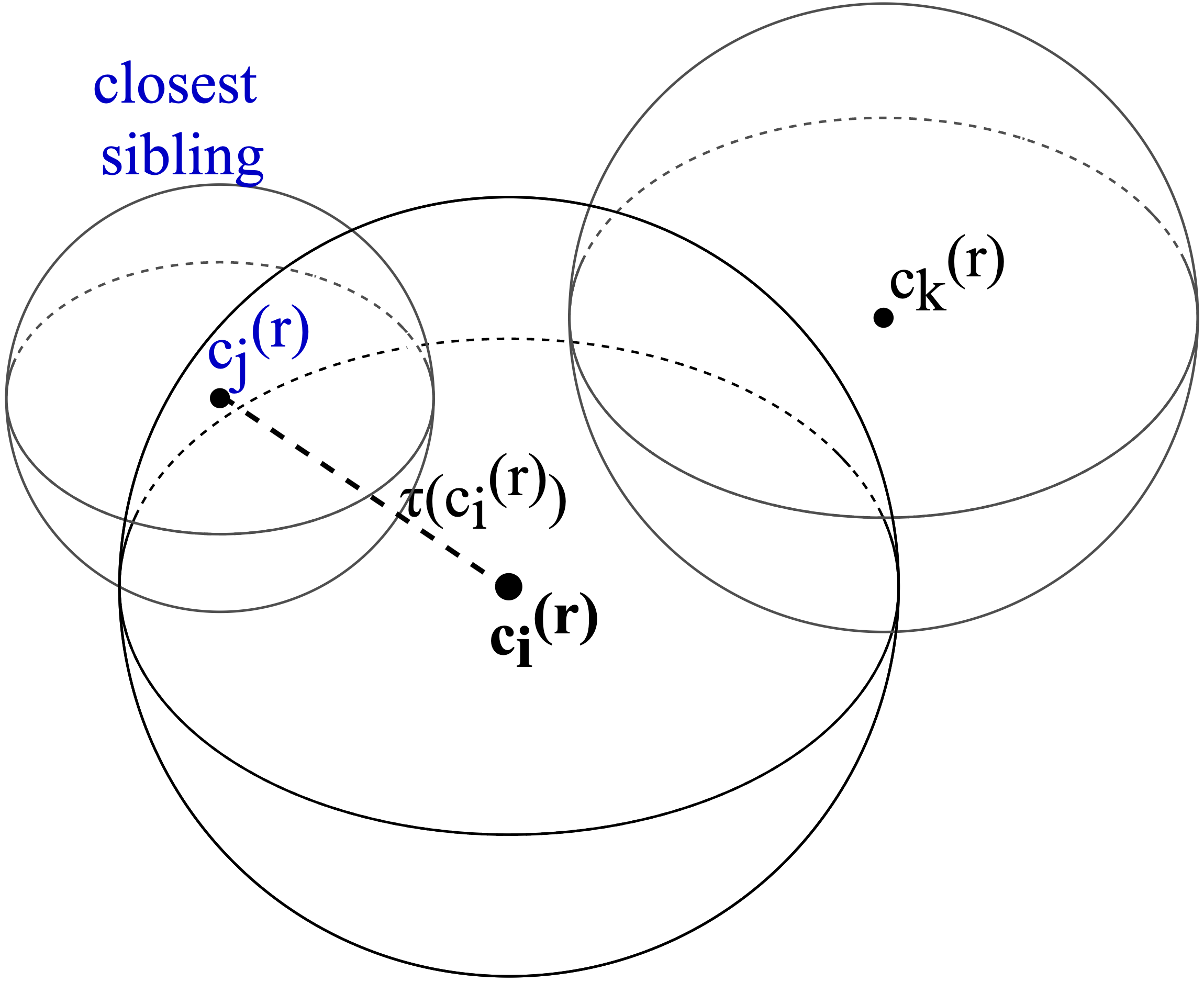} }}
\hfill
\caption{
(a) Extending the taxonomy by expanding the children of $c_i$ with "Other"  $c_{\text{other}}^{(i)}$. $\eta_i$ is the centroid of its sub-topic unit vectors. 
(b, c) Topic threshold $\tau(c_i^{(r)})$ for topic $c_i^{(r)}$ equal to (b) twice the distance of its furthest child when $\mathrm{ch}(c_i^{(r)})\ne \emptyset$ and (c) 
the distance of the nearest sibling, where $c_i^{(r)},c_j^{(r)},c_k^{(r)} \in \mathrm{ch}(c_r)$, when $\mathrm{ch}(c_i^{(r)})=\emptyset$. 
}
\label{tab:othercat_corethresh}
\end{figure*}

\subsubsection{Extending Taxonomy - Other Category}
\label{sec:extother}

A topic $c_i$ may also be unbalanced at a particular level $l$ if its children $c_j^{(i)} \in \mathrm{ch}(c_i)$ are unevenly distributed around it. Alternatively, it may just be incomplete due to partial enumeration.
Since topic representation updates are propagated up the taxonomy, such a topic would result in a bad representation and not only degrade the performance at that level, but propagate the imbalance upward.

One reason for this imbalance could be ``incompleteness" of $\mathcal{T}$ 
(i.e., the set of sub-topics is not fully enumerated).
Therefore, we introduce an ``Other'' topic $c_{\text{other}}^{(i)}$ as a subtopic of topic $c_i$
which accounts for its missing subtopics and balances out the hierarchy.
In particular, we calculate the
density of the original subtopics with respect to the main topic and 
then define $c_{\text{other}}^{(i)}$ such that it
pulls the density in the direction opposite to the centroid of the 
original subtopics in order to have a more even distribution of subtopics (see Figure \ref{tab:othercat_corethresh}).

The magnitude of $c_{\text{other}}^{(i)}$ (i.e., $\|c_{\text{other}}^{(i)}\|$) is approximated to be 
the magnitude of the centroid of the subtopics. 
Its representation is given by Equation \ref{eq:othercatequation} (see Appendix~\ref{sec:appendixOtherCat} for derivation). It depends on both its sibling subtopics as well as its parent topic.  We 
extend each $c_i$ in the taxonomy with $c_{\text{other}}^{(i)}$
in a bottom up manner from the leaf level and stop at $\rho$ (as we define the topics above $\rho$ to be complete). 

In our well-being taxonomy from Figure \ref{tab:example}, we can see why expansion via the addition of the ``Other" category is desirable.  Both the \textit{healthcare} and \textit{safety} subtrees are only partially enumerated.  Automatically expanding their subtopic sets avoids having to fully and painstakingly enumerate them.
\begin{equation}
\label{eq:othercatequation}
\begin{aligned}
    c_{\text{other}}^{(i)} \approx 
    c_i - \|c_{\text{other}}^{(i)}\| \sum_{c_j^{(i)} \in \mathrm{ch}(c_i)} \dfrac{c_j^{(i)} - c_i}{\|c_j^{(i)} - c_i\|}
\end{aligned}
\end{equation}

\subsubsection{Topic Threshold and Assignment}
\label{sec:topicthreshass}
To assign documents to topics, we initially perform distance based classification independently for each subtree at level $\rho$.
We perform classification only at level $\rho$ as there is a greater degree of confidence that the discovered documents actually belong to that topic
since we define all topics $c_i$ with $\lambda(c_i) < \rho$ as complete and balanced.

Let a \textit{root} topic $c_r \in C^{\rho-1}$ be a topic at level $\rho-1$. 
For each child topic
$c_i^{(r)} \in \mathrm{ch}(c_r)$
we fit a set of documents $d_k \in \delta_i^{(r)}$ belonging to it.
A document belongs to the topic if it is within a certain distance threshold (\textbf{topic threshold}) in the embedding space, thus partitioning the entire dataset.

$\tau(c_i^{(r)})$ is the learned threshold distance (or a maximum radius) for each topic $c_i^{(r)} \in \mathrm{ch}(c_r)$ when assigning documents to it.
It takes into account both the density of documents $d_k \in \delta_i^{(r)}$ around $c_i^{(r)}$ and the taxonomy $\mathcal{T}$ (siblings and children).
It is only defined for topics at level $\rho$. 
See Figures \ref{tab:othercat_corethresh}b, \ref{tab:othercat_corethresh}c and Appendix \ref{sec:appendixeccentricity} for derivation.

The initial \textit{assigned set} $\delta_i^{(r)}$, for a topic $c_i^{(r)}$ is:
\begin{equation}
\label{eq:coreassignment}
\begin{aligned}
    \delta_i^{(r)} = \{d_k\:|\:d_k\in \mathcal{D} \land \|d_k-c_i^{(r)}\|\le \tau(c_i^{(r)})\}
\end{aligned}
\end{equation}
If there are no documents within the topic threshold for a topic $c_i^{(r)}$,
we adapt to the document density by updating the topic's threshold to be at least equal to $\alpha \ge 1$ 
times the distance of the nearest document from $c_i^{(r)}$, if it is within twice the original threshold.

Finally, we
update the \textit{assigned sets} $\delta_i^{(r)}$
by re-performing the assignment. 
We control the \textit{degree of overlap} between the \textit{assigned sets} using an additional hyperparameter (see Appendix~\ref{sec:appendixeccentricity} for details).

\subsection{Cluster Assignment and Optimization}
\label{sec:shc}
Given the corpus $\mathcal{D}$, 
the updated taxonomy $\mathcal{T^{'}}$ (balanced, extended), 
the pivot level $\rho$ 
and the \textit{assigned sets},
we propose an EM-style algorithm iterating between the assignment of the document sets $\delta_i$ (\textbf{E-Step}) and recomputing the topic representations $c_i$ (\textbf{M-Step}). 
We maximize the expectation 
(i.e., likelihood of assigning a document to a topic assuming uniform probability) 
by minimizing the objective: 
\begin{equation}
\label{eq:objectivefn}
\begin{aligned}
    \mathcal{L} = \sum_{c_i\in\mathcal{T^{'}}} {
        \sum_{d_k\in\delta_i} {\lambda(c_i)\cdot\|d_k-c_i\|^2}
    }
\end{aligned}
\end{equation}

\paragraph{E-Step}
The topic thresholds are used to determine the \textit{assigned set} for each topic at $\rho$ (E-Step, line 6 in Algo. \ref{alg:shc}). 
Next, for each topic $c_i \in C^{l}$ at level $l$ with \textit{assigned set} $\delta_i$, we solve a K-Means formulation (by Voronoi Iterations \citep{lloyd1982least}) in a top-down manner from $l = \rho$ to $N$. 
The iterations are performed 
over the set $\delta_i$ to fit $k=|\mathrm{ch}(c_i)|$ clusters, with initial cluster centers $c_k^{(i)} \in \mathrm{ch}(c_i)$.

The obtained clusters correspond to the set of assigned documents $\delta_k^{(i)}$ for topic $c_k^{(i)}$ (E-top down, line 12). 
The process continues top-down for all sibling and successor topics until the leaves of $\mathcal{T^{'}}$.

\paragraph{M-Step}
As the cluster centers are updated (in E-top down), we set topic representations to 
corresponding cluster centroids (M-top down, line 9 in Algo. \ref{alg:shc}).
Now, as topic representations are a function of themselves and their children,
we compute bottom-up updates of topics (parents, up to level $\rho$) as discussed in §\ref{sec:taxorep} 
(M-bottom up, line 3).

Finally, as each topic has only one parent in the taxonomy, we complete the taxonomy by directly deriving the topics above the pivot level $\rho$ for each document set from their pivot level assignments.

\paragraph{Inference}
At inference time, we use our learned topic representations to assign documents to each topic.
We use Eq. \ref{eq:coreassignment} to obtain \textit{assigned sets} $\delta_i$ using the learned topic threshold $\tau(c_i)$ for each topic $c_i$ at level $\rho$.
Documents not within any pivot topic threshold are 
assigned to a \textit{None} category.
Then, 
each $\delta_i$ is split up among its children $c_j^{(i)} \in \mathrm{ch}(c_i)$ 
by assigning each document to its closest topic $c_j^{(i)}$ to obtain the sets $\delta_j^{(i)}$.
This process is repeated in a top-down manner to the leaf topics.

\paragraph{Complexity}
Each topic's representation is updated using Eq. \ref{eq:weightmeanupdaterep} which relies on its children and their $\mathrm{LES}$, 
with a complexity of $\mathcal{O} (|\delta|B^3)$ 
where $\delta$ is the set of documents assigned to it and $B$ is the hierarchy's maximal branching factor (see Appendix~\ref{sec:appendixLES}). Each topic is then extended using Eq. \ref{eq:othercatequation} with complexity $\mathcal{O} (B)$,
and its \textit{topic threshold} and \textit{assigned set} are obtained with Eq. \ref{eq:coreassignment} with complexity $\mathcal{O} (B+D)$ where $D$ is the size of the unlabeled corpus. 
The objective Eq. \ref{eq:objectivefn} 
identifies
a topic's cluster in $\mathcal{O} (D)$.
We do these at most $n$ times, once for each of our $n$ topics, until convergence.
In practice we found that convergence is achieved within $4$ iterations.
Overall, \textsc{HierSeed} scales linearly with taxonomy size $n$ and corpus size $D$.

\section{Experiment Details}

\subsection{Datasets}
We use
three publicly available datasets for evaluation: RCV1-V2~\citep{lewis2004rcv1}, NYTimes (NYT)~\citep{sandhaus2008new} and Web-of-Science (WOS)~\citep{kowsari2017hdltex}. 
RCV1-V2 and NYT are news categorization corpora while WOS includes categorization of published scientific paper abstracts.
All documents in WOS belong to a single leaf topic while documents in NYT and RCV1 may belong to multiple leaf/non-leaf topics. 
Data statistics are shown in Table \ref{data_statistics}. 

\begin{table}
\centering
\scalebox{0.8}{
\begin{tabular}{lrrr}
	 \hline
	 & \textbf{WOS} & \textbf{NYT} & \textbf{RCV1} \\
	 \hline
	 \hline
	 $|\mathcal{T}|$        & 141 & 166 & 103 \\
	 Height of $\mathcal{T}$                 & 2 & 8 & 4 \\
	 \hline
	 Training               & 37588 & 29179 & 23149 \\
	 \hspace{3mm}Seed ($|\mathcal{S}|$ with $|\mathcal{S}_i| \leq 4$) & 532 & 290 & 390 \\
	 \hspace{3mm}Fitting ($\mathcal{|D|}$)      & 37056 & 28889 & 22759 \\
	 Test                   & 9397 & 7292 & 781265 \\
	 \hline
\end{tabular}
}
\caption{\label{data_statistics}
Datasets statistics. 
$|\mathcal{T}|$ is the number of topics in the taxonomy. 
The training and test sizes are the main data splits. The seed $\mathcal{S}$ (labeled) and fitting $\mathcal{D}$ (unlabeled) sets are subsets of the training set.}
\end{table}



We use the benchmark train/test split for RCV1 and for NYT and WOS we randomly split the data. 
For each dataset, the training set is also split into the seed ($\mathcal{S}$) and fitting ($\mathcal{D}$) sets by randomly sampling a fixed number of documents $|\mathcal{S}_i|$ per topic $c_i$ as seeds. We only keep the labels for the much smaller seed sets and discard them for the fitting sets. 

\subsection{Metrics}
We evaluate our algorithm using $B^3$~\citep{bagga1998algorithms} and V-Measure~\citep{rosenberg2007v}. 
$B^3$ is a cluster evaluation metric that measures precision and recall of a topic distribution.
V-Measure is a conditional entropy metric which measures cluster homogeneity and completeness.
For both metrics, we average across all levels, weighted equally.
For a fair comparison, we report the same metrics for both our method and the baselines (instead of classification F1).

\begin{table*}[ht]
\centering
\scalebox{0.73}{
\begin{tabular}{ll|rrrrr|rrrrr}
    \hline
    & & \multicolumn{5}{c}{No Fitting} & \multicolumn{5}{c}{Fitting + Seed}\\
    & & $^{\blacklozenge}$HDLTex & $^{\blacklozenge}$HiAGM & \begin{tabular}[t]{@{}c@{}}$^{\blacklozenge}$HiLAP\\-RL\end{tabular} & $^{\blacklozenge}$HFT(M)& \begin{tabular}[t]{@{}l@{}}$^{\blacklozenge\bigstar}$\textbf{Unfit-}\\\textbf{HierSeed}\end{tabular} & $^{\spadesuit}$HClass & $^{\clubsuit}$hLDA & $^{\clubsuit}$TSNTM & $^{\spadesuit\bigstar}$JoSH &
    $^{\spadesuit\bigstar}$\textbf{HierSeed}\\
    \hline
    \multirow{3}{*}{$B^3$ F1} & \textbf{WOS} & 0.2349 & 0.4044 & 0.1858 & 0.3055 & 0.5414 & 0.6420 & 0.1972 & 0.2262 & 0.5940 & \textbf{ 0.7131}\\
    & \textbf{NYT} & 0.4840 & 0.4065 & 0.3451 & 0.4079 & 0.5040 & 0.5008 & 0.3811 & 0.3349 & 0.4692  & \textbf{0.6173}\\
    & \textbf{RCV1} & 0.4231 & 0.4567 & 0.4279 & 0.5041 & 0.4923 & 0.6034 & 0.3873 & 0.3726 & 0.5366  & \textbf{0.6546}\\
    \hline
    \multirow{3}{*}{V-Ms} & \textbf{WOS} & 0.0793 & 0.3467 & 0.1383 & 0.2729 & 0.5569 & 0.5984 & 0.1984 & 0.1916 & 0.5927  & \textbf{0.7661}\\
    & \textbf{NYT} & 0.2253 & 0.2264 & 0.1098 & 0.1562 & 0.4316 & 0.4461 & 0.1781 & 0.2052 & 0.4447  & \textbf{0.5340}\\
    & \textbf{RCV1} & 0.1614 & 0.2754 & 0.1602 & 0.3422 & 0.3952 & 0.4289 & 0.2336 & 0.3026 & 0.3591  & \textbf{0.4815}\\
    \hline
\end{tabular}
}
\caption{\label{results}
$B^3$ F1 and V-Measure (V-MS) on the WOS, NYT, RCV1 datasets. Methods are trained either using only the seed set (No Fitting) or the seed set and unlabeled fitting data (Fitting + Seed). 
Methods are
$^{\blacklozenge}$supervised classification, $^{\spadesuit}$weakly-supervised, and
$^{\clubsuit}$unsupervised. $^{\bigstar}$ indicates seeded. There are up to $4$ seeds per topic.
}
\end{table*}

\subsection{Hyperparameters and Baselines}
We use a pretrained RoBERTa-base \citep{liu2019roberta} model to obtain a $768$-dimensional embedding for each document
by taking the mean across the final hidden states of all tokens. Other hyperparameters are listed in Appendix \ref{sec:appendixHyperparam}.

We compare
\textsc{HierSeed} to hierarchical classification, clustering and topic modeling baselines. 
As a baseline, we also compare it to \textsc{HierSeed} trained 
without the unlabeled fitting data 
(\textbf{Unfit-\textsc{HierSeed}}). That is, Unfit-\textsc{HierSeed}
uses just the seed set for initial topic representations
followed by lines 3-5 of Algo.~\ref{alg:shc} to update them, and line 6-8 to assign documents.
We use only the labeled seed set ($\mathcal{S}$) for baselines requiring seeds or supervision and the unlabeled fitting set ($\mathcal{D}$) for unsupervised baselines.
We evaluate each model $5$ times and report their averages.

For weakly-supervised and unsupervised baselines we use:
\textbf{hLDA}~\citep{griffiths2003hierarchical} -- an unsupervised non-parametric hierarchical topic model and \textbf{TSNTM}~\citep{isonuma2020tree} -- an unsupervised generative neural topic model, trained on the unlabeled fitting set;
\textbf{HClass}~\citep{meng2019weakly} -- a hierarchical classification model that uses
keywords from the seed set for pretraining and the unlabeled fitting set for self-training; and \textbf{JoSH}~\citep{meng2020hierarchical} -- a generative hierarchical topic mining model that uses the taxonomy for supervision, trained on the unlabeled fitting set. 
JoSH is the only seeded hierarchical method used for comparison. 

We additionally compare to a number of supervised approaches:
\textbf{HDLTex}~\citep{kowsari2017hdltex} -- a hierarchical classification model 
trained with the labeled seed set, \textbf{HiAGM}~\citep{zhou2020hierarchy} --  a hierarchical text classification model trained with the labeled seeds,
\textbf{HiLAP-RL}~\citep{huang2021hierarchy} -- a hierarchical classification technique trained with reinforcement learning,
and \textbf{HFT}~\citep{shimura2018hft} -- a hierarchical CNN-based text classifier, trained
on the seed set.
Further details about the baselines can be found in Appendix \ref{sec:appendixBaselines}.

\section{Results and Analysis}
\label{sec:experiments}

\subsection{Main Results}
Results are shown in Table~\ref{results}. Our method \textsc{HierSeed}, trained with labeled seeds and unlabeled fitting sets, outperforms all 
baselines
on the SHC task 
when restricted to the same training data.
Its best score is for the WOS corpus, which we hypothesize is due to its simpler taxonomy compared to NYT and RCV1. Additionally, even without fitting on the unlabeled data our method (Unfit-\textsc{HierSeed}) demonstrates strong performance, outperforming most baselines. In particular, Unfit-\textsc{HierSeed} (which does not use fitting data) is only outperformed by the two baselines that \textit{do} use the fitting data (HClass and JoSH). In fact, \textsc{HierSeed} \textit{with} fitting data outperforms both these methods by a large margin across all corpora and metrics. This shows the effectiveness of using a small labeled seed set to fit to a taxonomy.

Although the Unfit-\textsc{HierSeed} 
outperforms most baselines, there is still a large 
performance drop
compared to 
\textsc{HierSeed} 
(with fitting on unlabeled data).
Since Unfit-\textsc{HierSeed} does not use the unlabeled data (it stops the training after the E-Step, line 6 Algo.~\ref{alg:shc}, of the first iteration) it does not estimate the LES 
or update the \textit{topic thresholds}. Therefore,
it may inadvertently kill a branch of the hierarchy and so is limited in its ability to fit to the data.
The performance drop
shows the importance of
LES in 
computing better topic representations and fitting to the 
provided 
taxonomy.

\begin{figure}[tp]
\centering
\includegraphics[width=\linewidth]{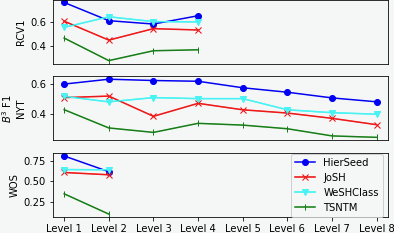}
\caption{Performance ($B^3$ F1) at each level of the hierarchy for all three evaluation datasets and the best performing model from each category.}
\label{tab:levelsF1}
\end{figure}

\begin{table}
\centering
\resizebox{\linewidth}{!}{%
\begin{tabular}{l|rrr|rrr}
    \hline
    &
    \multicolumn{3}{c}{$B^3$ F1} &
    \multicolumn{3}{c}{V-Measure} \\
    & \textbf{WOS} & \textbf{NYT} & \textbf{RCV1} & \textbf{WOS} & \textbf{NYT} & \textbf{RCV1}\\
    \hline
    \hline
    HDLTex & 0.7310 & 0.6483 & 0.6163 & 0.6446 & \textbf{0.6820} & 0.4155 \\
    HiAGM & \textbf{0.7528} & 0.5802 & 0.6363 & 0.6531 & 0.5904 & 0.3897 \\
    HiLAP-RL & 0.5641 & 0.5116 & 0.6438 & 0.5430 & 0.3083 & 0.4037 \\
    HFT(M) & 0.6943 & \textbf{0.7130} & \textbf{0.6902} & 0.7372 & 0.6133 & \textbf{0.5472} \\
    \hline
    \textbf{\textsc{HierSeed}} & 0.7131 & 0.6173 & 0.6546 & \textbf{0.7661} & 0.5340 & 0.4815 \\
    \hline
\end{tabular}%
}
\caption{\label{baselinefulltrainresults}
Results of training the classification baselines with the full labeled training set (with 20\% validation), compared to \textsc{HierSeed} trained only using the labeled seed and unlabeled fitting sets, using 4 seeds per topic.
}
\end{table}


Comparing baselines, we see that the unsupervised methods (hLDA and TSNTM) perform poorly compared to the (weakly-)supervised classification methods.
Although unsupervised approaches are good for discovering latent hierarchies, they aren't capable of generating topics similar to a predefined structure. Additionally, most supervised classification methods still perform poorly compared to \textsc{HierSeed} since these models do not make use of unlabeled data and have only a small set of seed examples for supervision.

We also examine the affect of taxonomy depth on performance.
Figure \ref{tab:levelsF1} shows $B^3$ F1 at each level.
Although, performance degrades at deeper levels of the hierarchy,
our method is consistently better than the others 
at deeper levels 
highlighting
our system's ability to learn better topic representations.

We note that although supervised classification baselines trained on the entire labeled training set outperform our method (Table~\ref{baselinefulltrainresults}), we achieve competitive results using a substantially smaller labeled set. 
The strength of our method is in the weakly-supervised nature of the training procedure and it is therefore better suited to real-world data-scarce settings than fully supervised approaches.

Thus, we see the advantages of weakly-supervised approaches, and especially \textsc{HierSeed}, 
which can both 
adhere to a predefined structure 
(i.e., a labeled taxonomy),
and
make good use of the much easier to obtain unlabeled fitting set.

\subsection{System Analysis}
We conduct analysis on the number of seeds per topic and the document representation method. 

\begin{table}
\centering
\resizebox{\linewidth}{!}{%
\begin{tabular}{l|rrr|rrr}
    \hline
    &
    \multicolumn{3}{c}{$B^3$ F1} &
    \multicolumn{3}{c}{V-Measure} \\
    & \textbf{WOS} & \textbf{NYT} & \textbf{RCV1} & \textbf{WOS} & \textbf{NYT} & \textbf{RCV1}\\
    \hline
    \hline
    \# seed = 2 & 0.6701 & 0.6078 & 0.6654 & 0.7395 & 0.5021 & 0.5073 \\
    \# seed = 4 & 0.7131 & 0.6173 & 0.6546 & 0.7661 & 0.5340 & 0.4815 \\
    \# seed = 6 & 0.7284 & 0.6388 & 0.6870 & 0.7762 & 0.5452 & 0.5225 \\
    \# seed = 8 & \textbf{0.7288} & \textbf{0.6410} & \textbf{0.6926} 
    & \textbf{0.7892} & \textbf{0.5538} & \textbf{0.5318}\\
    \hline
\end{tabular}%
}
\caption{\label{experimentresults}
\textsc{HierSeed} with different numbers of seeds per topic in the taxonomy. The models are trained on their respective seeds and fitted on the fitting sets.
}
\end{table}


\paragraph{Number of Seed Examples}
We experiment with using different numbers of seeds for each topic in the taxonomy (see  
Table~\ref{experimentresults}). We see a general increasing trend in the performance with an increasing number of seeds as the topics become more representative. However, the trend approaches saturation when going from $6$ to $8$ showcasing how little annotated data is required, to be effective.

\begin{table}
\centering
\resizebox{\linewidth}{!}{%
\begin{tabular}{l|rrr|rrr}
    \hline
    &
    \multicolumn{3}{c}{$B^3$ F1} &
    \multicolumn{3}{c}{V-Measure} \\
    & \textbf{WOS} & \textbf{NYT} & \textbf{RCV1} & \textbf{WOS} & \textbf{NYT} & \textbf{RCV1}\\
    \hline
    \hline
    RoBERTa & \textbf{0.7131} & 0.6173 & 0.6546 & \textbf{0.7661} & \textbf{0.5340} & 0.4815 \\
    GloVe-300d & 0.7039 & 0.6191 & \textbf{0.6692} & 0.7524 & 0.5272 & \textbf{0.5188} \\
    fastText & 0.7125 & \textbf{0.6202} & 0.6524 & 0.7574 & 0.5033 & 0.4594 \\
    \hline
\end{tabular}%
}
\caption{\label{embeddingexperiments}
Performance of \textsc{HierSeed} using different embeddings, trained using the labeled seed and unlabeled fitting sets, using 4 seeds per topic.
}
\end{table}


\paragraph{Embeddings for Document Representations}
To test \textsc{HierSeed}'s dependence on the nature (contextual vs. static) and quality of embeddings, 
we switch out the document embeddings.
In Table \ref{embeddingexperiments}, we compare the performance of \textsc{HierSeed} using RoBERTa (used in all other experiments),
300-dimensional GloVe~\citep{pennington2014glove},
and fastText skip-gram ~\citep{joulin2016bag} 
while using $4$ seeds per topic.
The document embeddings are obtained by taking the mean of all tokens.

We see that \textsc{HierSeed} performs equally well for each embedding, with GloVe performing better on RCV1, and fastText on NYT. 
However the performance differences are small, and consistently better than the baselines,
showing that \textsc{HierSeed} is embedding-agnostic and is able to identify a good representation for the taxonomy
regardless.

\subsection{Error Analysis}
An analysis of \textsc{HierSeed}'s hierarchical assignments
highlights some important shortcomings and modes of failures. 
First,
mistakes are more likely at the pivot level than in subsequent levels. 
This is intuitive since
taxonomies get more specific (i.e., easier to fit to) down the hierarchy
and \textsc{HierSeed} assigns documents top-down from the pivot level. 

In addition, 
a small set of seeds for a topic may not cover all subtopics, especially if there are many semantically diverse subtopics
(e.g., `Vaccines', `Enzymes',
and `Cancer' are diverse subtopics of  `Molecular Biology').
Furthermore,
if \textit{topics} are semantically similar (e.g., `consumer finance' vs. `government finance'), then the seed documents (and topic representations) may also be similar making it difficult to distinguish between the topics.
Additionally, errors come from a lack of domain specific embeddings or 
informative
document representations.
For example, corpus specific artifacts such as jargon, equations, numeric data (e.g., in WOS) and tables and figures (e.g., in NYT)
can lead to
uninformative document embeddings that result in incorrect topic assignment.

Finally, our method assumes an incomplete taxonomy (i.e., always adds an Other category) and therefore cannot distinguish between None and Other below the pivot level.
For example, 
an author biography $\alpha$ from NYT 
is assigned as "Feature (level 1 - pivot level) - Books (level 2) - Other (level 3)" instead of "Feature - Books - None" (in a taxonomy consisting of just book genres).
This is because, once $\alpha$ is assigned to the topic ``Feature'', it can no longer be assigned None at the following levels. However, in general we find our assumption of incompleteness is valid.

\section{Conclusion and Future Work}
In this paper, we formalize the task of Seeded Hierarchical Clustering: 
fitting a large, unlabeled corpus to a user-defined taxonomy (that may be unbalanced or incomplete)
using only a small number of labeled examples.
We propose a novel, discriminative weakly supervised algorithm, \textsc{HierSeed}, for it
which outperforms both unsupervised and (weakly) supervised state-of-the-art techniques 
on three real-world datasets from different domains.

In the future, we aim to jointly learn and fine-tune task specific embeddings, develop a generative variant of \textsc{HierSeed} and explore non-Euclidean representation spaces.

\bibliography{anthology,custom}
\bibliographystyle{acl_natbib}

\clearpage
\appendix

\section{Derivation of Other Category}
\label{sec:appendixOtherCat}

For a topic $c_i$ and its children $\mathrm{ch}(c_i)$, we introduce an "Other" category $c_{\text{other}}^{(i)}$ at its subtopic level to "complete" the set and denote it by $\mathrm{ch}^{\prime}(c_i)$ such that $\mathrm{ch}^{\prime}(c_i) = \mathrm{ch}(c_i) \cup \{c_{\text{other}}^{(i)}\}$. 
To do so, we introduce the concept of \textit{degree of imbalance}.

\textbf{Degree of imbalance} - $\eta$ measures the distance of the centroid of the sub-topic representation vectors from the main topic representation, if these sub-topic were points on a unit-sphere around the main topic. A balanced hierarchy has $\eta=0$ and an unbalanced hierarchy has $\eta$ closer (never equal) to 1.

Let, $\widehat{c}_j^{(i)}$ be a unit directional vector from the topic $c_i$ to a sub-topic $c_j^{(i)} \in \mathrm{ch}(c_i)$. Then,
\begin{equation}
\label{eq:unitdirvec}
\begin{aligned}
    \widehat{c}_j^{(i)} = \dfrac{c_j^{(i)} - c_i}{\|c_j^{(i)} - c_i\|}
\end{aligned}
\end{equation}

The \textit{degree of imbalance} $\eta_i$ for $c_i$ is given by the centroid of all $\widehat{c}_j^{(i)}$ as:
\begin{equation}
\label{eq:degreeimbalance}
\begin{aligned}
    \eta_i = \dfrac{1}{|\mathrm{ch}(c_i)|} \sum_{c_j^{(i)} \in \mathrm{ch}(c_i)} \widehat{c}_j^{(i)}
\end{aligned}
\end{equation}

We want to decrease the \textit{degree of imbalance} of the augmented sub-topic set $\mathrm{ch}^{\prime}(c_i)$ such that, 
\begin{equation*}
\begin{aligned}
    \eta^{\prime}_i \approx 0
\end{aligned}
\end{equation*}
\begin{equation*}
\begin{aligned}
    \implies
    &\dfrac{1}{|\mathrm{ch}^{\prime}(c_i)|} \sum_{c_j^{(i)} \in \mathrm{ch}^{\prime}(c_i)} \widehat{c}_j^{(i)} \\
    &= \dfrac{1}{|\mathrm{ch}(c_i)|+1} \sum_{c_j^{(i)} \in \mathrm{ch}(c_i)} \widehat{c}_j^{(i)}
    + \widehat{c}_{\text{other}}^{(i)} \\
    &\approx 0
\end{aligned}
\end{equation*}
Therefore,
\begin{equation}
\label{eq:otherunitdirvector}
\begin{aligned}
    \widehat{c}_{\text{other}}^{(i)} \approx - |\mathrm{ch}(c_i)| \cdot \eta_i
\end{aligned}
\end{equation}

Solving for $c_{\text{other}}^{(i)}$ using Equations \ref{eq:unitdirvec} and \ref{eq:otherunitdirvector}:
\begin{equation*}
\begin{aligned}
    c_{\text{other}}^{(i)} \approx 
    -|\mathrm{ch}(c_i)|\cdot\|c_{\text{other}}^{(i)} - c_i\| \cdot \eta_i 
    + C_i
\end{aligned}
\end{equation*}

We can set the magnitude of the "Other" vector to approximately be:
\begin{equation*}
\begin{aligned}
    \|c_{\text{other}}^{(i)} - c_i\| 
    \approx \left\| \dfrac{1}{|\mathrm{ch}(c_i)|} \sum_{c_j^{(i)} \in \mathrm{ch}(c_i)} (c_j^{(i)}-c_i) \right\|
\end{aligned}
\end{equation*}
i.e. distance of the centroid of the sub-topics $c_j^{(i)} \in \mathrm{ch}(c_i)$ from the topic $c_i$.

Therefore, on substituting the values we get,
\begin{equation}
\begin{aligned}
    c_{\text{other}}^{(i)} \approx 
    -\eta_i \cdot \left\| \sum_{c_j^{(i)} \in \mathrm{ch}(c_i)} (c_j^{(i)}-C_i) \right\| 
    + c_i
\end{aligned}
\end{equation}

This gives us the "Other" category to augment the sub-topics set $\mathrm{ch}(c_i)$ (Figure \ref{tab:othercat_corethresh}a shows the configurations in embedding space). 
We apply this in a bottom up manner starting from level $N$ to $\rho$ (pivot level) as we assume that the topics above it are complete. 
From the equation we see that the representation of the "Other" category depends on representations of sibling sub-topics as well as the parent topic.

To compute $c_{\text{other}}^{(k)}$ for $\mathrm{ch}(c_k)$ such that $\lambda(c_k)=\rho-1$ (i.e. finding the "Other" category for topics at pivot level), we do not readily have $c_k$. 
Here, we first set $c_k= \mathrm{LES}\big(\mathrm{ch}(c_k)\big)$, and then compute $c_{\text{other}}^{(k)}$.

\section{Approximating LES}
\label{sec:appendixLES}

\begin{figure}[tp]
\centering
\includegraphics[width=0.63\linewidth]{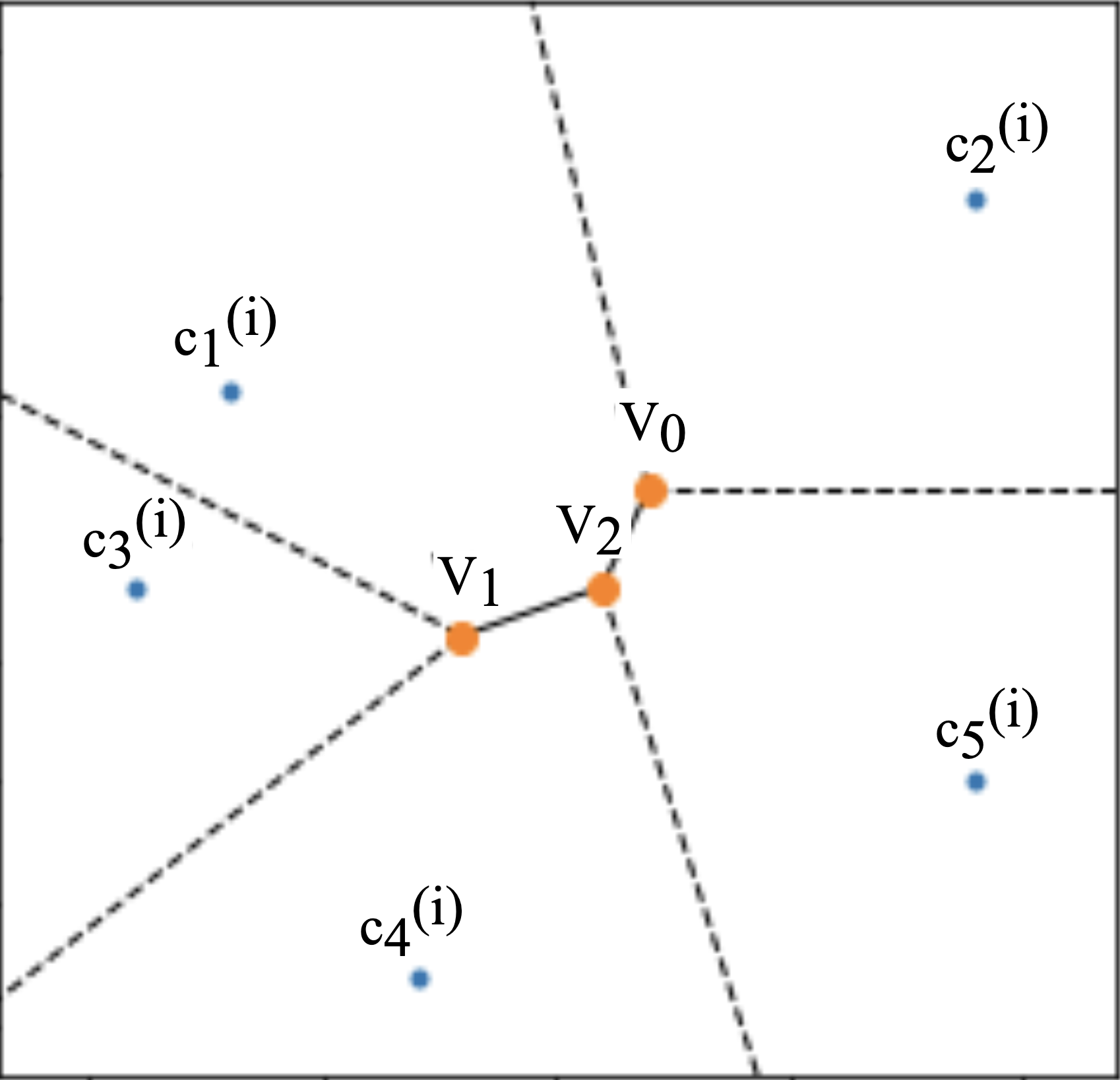}
\caption{Voronoi diagram of a set of points $\mathcal{P} = \mathrm{ch}(c_i)$. The Voronoi vertices are shown with $V_i$ in the diagram. There are three such vertices here.}
\label{tab:voronoi}
\end{figure}

\begin{figure*}[tp]
\centering
\hfill
\subfloat[\centering Sub-topics and data]{{\includegraphics[width=5cm]{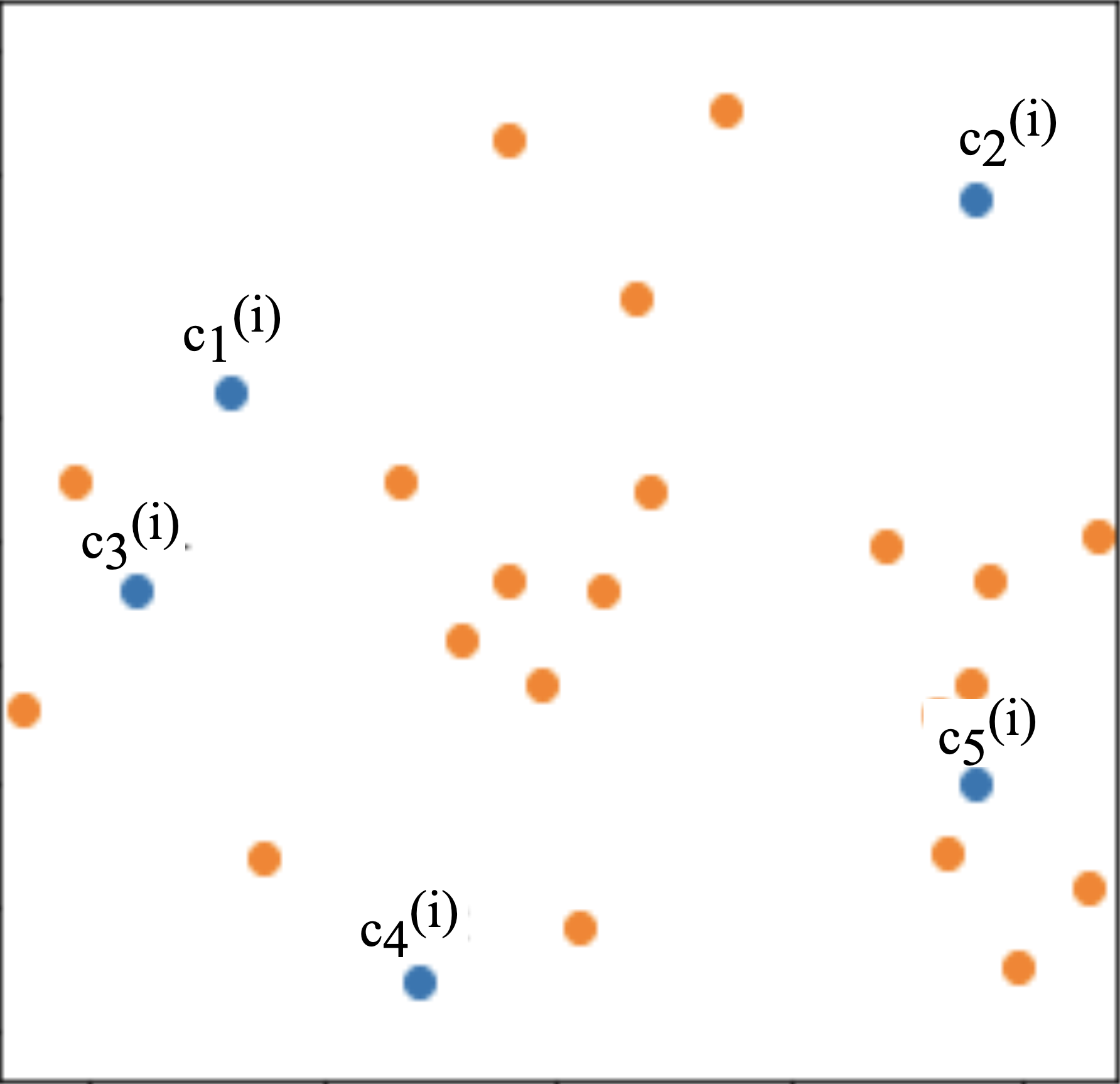} }}
\hfill
\subfloat[\centering Approx Voronoi vertices]{{\includegraphics[width=5cm]{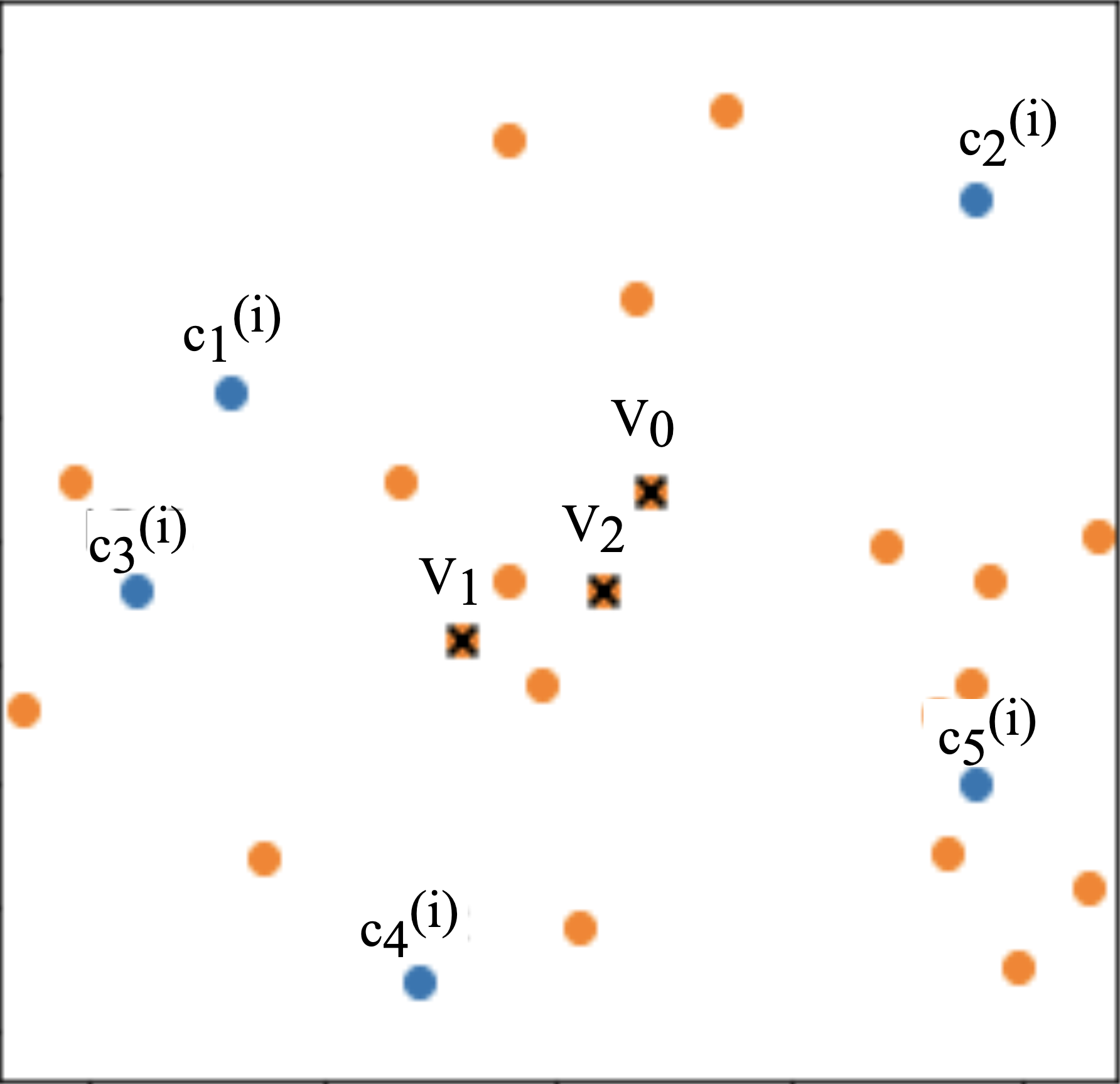} }}
\hfill
\subfloat[\centering Delaunay triangle Spheres]{{\includegraphics[width=5cm]{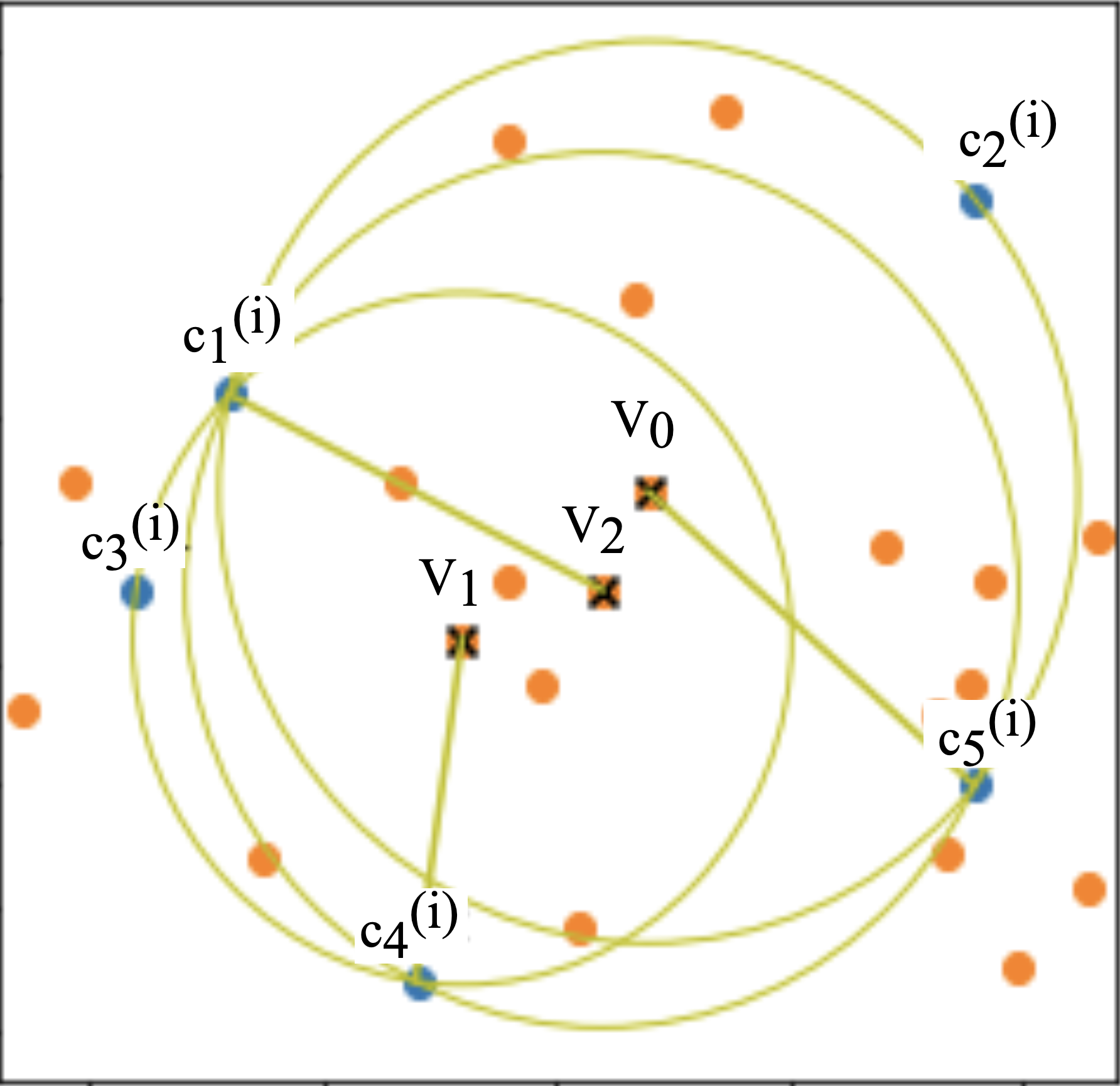} }}
\hfill
\caption{Approximating set of Voronoi vertices $\widehat{V}_i$ from set of sub-topic points $\mathcal{P} = \mathrm{ch}(c_i)$ and data points $d_k \in \delta_i$ for a topic $c_i$. (a) data points colored orange and $\mathcal{P}$ colored blue. (b) Approx Voronoi vertices $v_k \in \widehat{V}_i$ labeled with black $\times$. $\widehat{V}_i \subset \delta_i$. (c) Circumcircles (spheres) drawn with $v_k$ as center and $p \in \mathcal{P}$ on circumference.}
\label{tab:approxvoronoi}
\end{figure*}

The \textit{LES} for a set of points is found by constructing the Voronoi diagram (Figure \ref{tab:voronoi}) which divides a space such that all points within a region are closest to a point $p \in \mathcal{P}$, than to any other point $p^{\prime} \in \mathcal{P}$.
The center of a \textit{LES} is always either a Voronoi vertex or is the point of intersection of a Voronoi edge and the convex hull \citep{schuster2008largest}. 

In our case, the set of points $\mathcal{P} = \mathrm{ch}(c_i)$ for each topic $c_i \in \mathbb{R}^d$.
We know for most practical purposes $d \gg |\mathcal{P}|$ and this makes the problem intractable with no solution. 
Thus, we propose an approximate form of Voronoi decomposition.


Ideally, the sub-topics $\mathcal{P}$ and the assigned subset of documents $\delta_i$ for topic $c_i$, would be distributed such that a distance based formulation such as K-Means would converge with centers equal to $\mathcal{P}$, and the decision boundaries would construct the Voronoi diagram. 
We know that the documents $d_k \in \delta_i$ represent a subset of the set of infinite spatial points $\Phi_i$ around $c_i$, i.e., $\delta_i \subset \Phi_i$. 
Also, the set of Voronoi vertices $v_k \in V_i$ for topic $c_i$ satisfies $V_i \subset \Phi_i$, and bounded by $\delta_i$.

The duality of Voronoi diagrams and Delaunay triangulation states that Voronoi vertices are the circumcenters of Delaunay triangles, where the vertices of the triangles are from $\mathcal{P}$. 
That is, for every $v_k \in V_i$ there are three points $p_1, p_2, p_3 \in \mathcal{P}$:
\begin{equation*}
\begin{aligned}
    \|p_1 - v_k \| = \|p_2 - v_k \| = \|p_3 - v_k \|
\end{aligned}
\end{equation*}
Since $V_i \subset \Phi_i$ one can iterate through all points in $\Phi_i$ to find all $v_k$ that satisfy this equality. 
However, since $\Phi_i$ is infinite, we approximate $V_i$ from $\delta_i$ instead (see Figure \ref{tab:approxvoronoi}).

The approximate set of Voronoi vertices $\widehat{V}_i$ such that, $\widehat{V}_i \subset \delta_i$, and for all combination of three points $p_1, p_2, p_3 \in \mathcal{P}$ is given by:
\begin{equation}
\label{eq:approxvoronoi}
\begin{aligned}
    \widehat{V}_i = \{ &d_{k'} \:|\: 
                        d_{k'} \in \delta_i \land \\
                        &\|p_1 - d_{k'} \| \approx \|p_2 - d_{k'} \| \approx 
                        \|p_3 - d_{k'} \|
                    \}
\end{aligned}
\end{equation}

Since there are $\mathcal{O} (|\mathcal{P}|^3)$ such combinations and we iterate through $\delta_i$, the time complexity of this approximate algorithm is $\mathcal{O} (|\mathcal{P}|^3 |\delta_i|)$ for each topic.


This approximation is good if the set $\delta_i$ is dense. 
Additionally, the error threshold for the equality in Equation \ref{eq:approxvoronoi} may be determined based on the statistics (mean and standard deviation) of all errors.


Finally, the $\mathrm{LES}\big(\mathrm{ch}(c_i)\big)$ is the center of the sphere with the largest radius (Figure \ref{tab:approxvoronoi}c).

\section{Topic Threshold and Assigned Set Overlap}
\label{sec:appendixeccentricity}

\begin{figure}[tp]
\centering
\includegraphics[width=4.5cm]{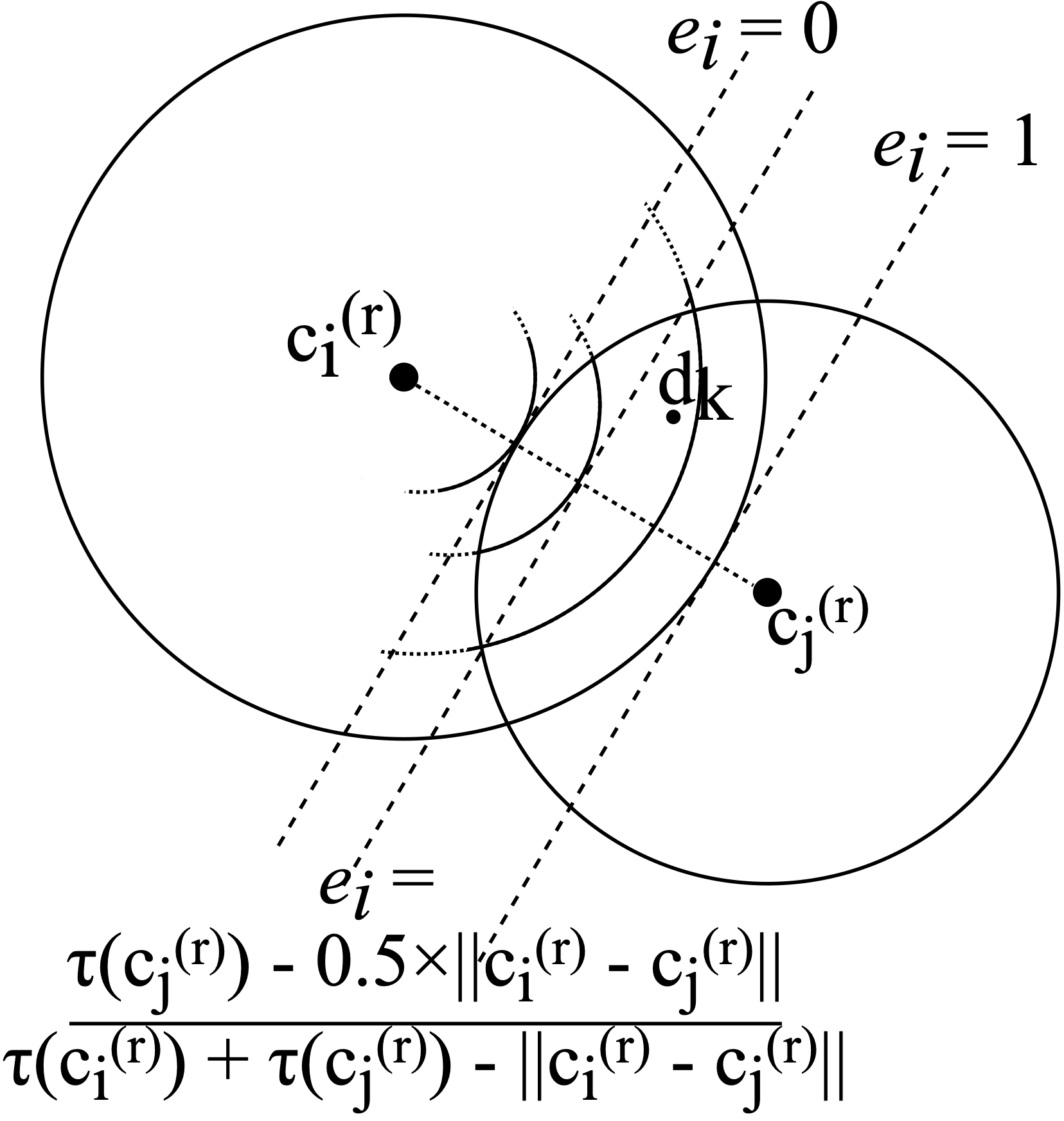}
\caption{Three eccentricity settings $e_i$ for the topic $C_i$ w.r.t $C_j$ to resolve \textit{assigned set} overlap.
}
\label{tab:coreoverlap}
\end{figure}

\textbf{Topic Threshold} - 
For each $c_i^{(r)} \in \mathrm{ch}(c_r)$, if $\mathrm{ch}(c_i^{(r)}) \ne \emptyset$, the topic threshold $\tau(c_i^{(r)})$ is given by Equation \ref{eq:corethreshold} (twice the distance of the furthest child), else, by Equation \ref{eq:corethreshold2} (distance of the nearest sibling).
\begin{equation}
\label{eq:corethreshold}
\begin{aligned}
    \tau(c_i^{(r)}) = 
    2\times \displaystyle\max_{c_j^{(i)} \in\: \mathrm{ch}(c_i^{(r)})} {\|c_i^{(r)}-c_j^{(i)}\|}
\end{aligned}
\end{equation}
\vspace{-20pt}
\begin{equation}
\label{eq:corethreshold2}
\begin{aligned}
    \tau(c_i^{(r)}) = 
    \displaystyle\min_{c_j^{(r)} \:\in\: \mathrm{ch}(c_r)-\{c_i^{(r)}\}} {\|c_i^{(r)}-c_j^{(r)}\|}
\end{aligned}
\end{equation}

\textbf{Eccentricity} - there may be overlap between the \textit{assigned sets} $\delta_i$ for each topic $c_i$. Thus, to control the degree of overlap between these sets, we introduce a parameter called \textit{eccentricity} $e_i$, for each of these topics $c_i$, such that $e_i \in [0,1]$ (see Figure \ref{tab:coreoverlap}).

For a topic $c_i^{(r)}\in \mathrm{ch}(c_r)$ (for a topic $c_r \in C^{\rho-1}$) and each sibling topic $c_j^{(r)}\in \mathrm{ch}(c_r)-\{c_i^{(r)}\}$ such that $\delta_i \cap \delta_j \ne \emptyset$, consider a document $d_k \in \delta_i \cap \delta_j$. We keep $d_k$ in $\delta_i$ if:
\begin{equation}
\label{eq:seteccentricity}
\begin{aligned}
    \|d_k-c_i^{(r)}\| \le{} &\|c_i^{(r)}-c_j^{(r)}\| - \tau(c_j^{(r)}) + \\
                      &e_i\Big(\tau(c_i^{(r)})+\tau(c_j^{(r)}) \\
                               &-\|c_i^{(r)}-c_j^{(r)}\|\Big)
\end{aligned}
\end{equation}

The most common setting for $e_i$ is:
\begin{equation*}
\begin{aligned}
    e_i = \dfrac{\tau(c_j^{(r)})-0.5\times\|c_i^{(r)}-c_j^{(r)}\|}{\tau(c_i^{(r)})+\tau(c_j^{(r)})-\|c_i^{(r)}-c_j^{(r)}\|}
\end{aligned}
\end{equation*}
which assigns $d_k$ to $c_i^{(r)}$ if it is nearer to it, compared to $c_j^{(r)}$. 
This is done for all topics $c_i^{(r)}\in \mathrm{ch}(c_r)$ to obtain the final \textit{assigned sets} $\delta_i$.

\section{Hyperparameter Settings}
\label{sec:appendixHyperparam}

The topic representations (§\ref{sec:taxorep}) are computed by choosing a pivot level $\rho=2$ for all three datasets. 

The weights for the weighted-mean (Equation \ref{eq:weightmeanupdaterep}) are set to $\frac{1}{|\mathrm{ch}(c_i)|}$ for all terms in $\mathrm{ch}(c_i)\}$, and the weight of $\mathrm{LES}(\mathrm{ch}(c_i))$ is set to 4 for WOS, RCV1, and to 0 for the NYT dataset. 

The value of $\alpha$ is set to 1.1 while recomputing the topic thresholds in case originally discovered \textit{assigned sets} are empty.

For each topic pair $c_i, c_j \in \mathcal{T}$, we set the eccentricity to be:
$$
    e_i = \frac{\tau(c_j)-0.5\times\|c_i-c_j\|}{\tau(c_i)+\tau(c_j)-\|c_i-c_j\|}
$$
(in Equation \ref{eq:seteccentricity}). 

Finally, we allow the algorithm to extend the taxonomy at all levels, with the "Other" category, for all three datasets. 

We also report our performance without fitting on the unlabeled fitting set, utilizing just the seed set for training. 
In Table \ref{results}, we set $c=4$ for all three datasets, and we alleviate the randomness by repeating the seed sampling process 5 times and reporting the average metrics.

\section{Baselines}
\label{sec:appendixBaselines}

We compare our with multiple baselines spanning unsupervised/seed-guided hierarchical topic models, unsupervised/seed-guided text embedding models, and weakly-supervised/supervised hierarchical text classification models.

\begin{itemize}[leftmargin=*]
    \setlength{\itemsep}{0pt}
    \setlength{\parskip}{0pt}
    \setlength{\parsep}{0pt}
    \item hLDA \citep{griffiths2003hierarchical}: a non-parametric hierarchical topic model based on the nested Chinese restaurant process with collapsed Gibbs sampling, which assumes documents are generated from a word distribution of a path of topics. Since it is unsupervised we use the training set (seed+fitting set) without any labels, and obtain the dynamic topic clusters.
    \item TSNTM \citep{isonuma2020tree}: a generative neural topic model which uses VAE inference to detect topic hierarchies in documents. Being unsupervised, we treat it as a clustering method which decides the hierarchical clusters dynamically, by fitting on the unlabeled training set.
    \item JoSH \citep{meng2020hierarchical}: a weakly-supervised generative hierarchical topic mining model which uses a joint tree and text embedding method to simultaneously model the category tree structure and the corpus generative process in the spherical space. We use this as a text classifier, trained on the unlabeled training set using the topic taxonomy as the supervision.
    \item WeSHClass \citep{meng2019weakly}: a weakly-supervised hierarchical classification model which leverages the provided keywords of each topic to generate a set of pseudo documents for pretraining and then self-trains on unlabeled data, using Word2Vec as embeddings. We use the keywords from the seed set for pretraining and the unlabeled fitting set for self-training.
    \item HDLTex \citep{kowsari2017hdltex}: a supervised method that combines multiple deep learning approaches in a top-down manner to produce hierarchical classification, by creating specialized architectures for each level of the hierarchy. Of the multiple variants presented, we use the RNN-RNN combination and train the model with the labeled seed set.
    \item HiAGM \citep{zhou2020hierarchy}:  an end-to-end hierarchical structure-aware global model that learns hierarchy-aware label and structure embeddings, formulated as a directed graph, which is then fused with text features to produce hierarchical text classification. We use the HiAGM-TP model and the GCN structure encoder, and train with the labeled seeds.
    \item HiLAP-RL \citep{huang2021hierarchy}: a top-down reinforcement learning based approach to hierarchical classification, where modeled as a Markov decision process and learns a label assignment policy. We use HiLAP with bow-CNN as the policy model.
    \item HFT \citep{shimura2018hft}: a hierarchical CNN fine-tuning based approach for text classification where  the model learns a classifier for the upper class labels, and uses transfer learning for the lower classes, thereby directly utilizing the parental/children dependency between adjacent levels. We train the HFT-CNN model using the recommended scoring function (MSF), on the seed set.

\end{itemize}

\end{document}